\documentclass{article}

\usepackage[dblblindworkshop, final]{neurips_2026}
\usepackage{svg}
\usepackage[utf8]{inputenc}
\usepackage[T1]{fontenc}
\usepackage{hyperref}
\hypersetup{hidelinks}
\usepackage{url}
\usepackage{booktabs}
\usepackage{amsmath,amsfonts}
\usepackage{nicefrac}
\usepackage{microtype}
\usepackage{xcolor}
\usepackage{graphicx}
\usepackage{multirow}
\usepackage{array}
\usepackage{adjustbox}
\usepackage{float}
\usepackage{amssymb}
\usepackage{caption}
\usepackage{placeins}
\usepackage{stfloats}
\usepackage[table]{xcolor}
\definecolor{colBlueLight}{RGB}{210,225,245}
\definecolor{colGreenLight}{RGB}{210,240,220}
\usepackage[ruled,vlined]{algorithm2e}
\usepackage{wrapfig}
\usepackage{graphicx}
\usepackage{subcaption}
\SetAlgoCaptionSeparator{.}
\RestyleAlgo{ruled}
\providecommand{\orcidlink}[1]{}
\providecommand{\keywords}[1]{}
\title{DeCO: Discriminative Evidence Composition for Fine-Grained Dataset Distillation}

\workshoptitle{Foundations of Efficient Deep Learning}

\author{
\textbf{Chuixuan Fan}$^{1,\dagger}$ \quad
\textbf{Guang Li}$^{2,\dagger,*}$ \quad
\textbf{Shijie Wang}$^{3}$ \quad
\textbf{Dongzhan Zhou}$^{4}$ \\
\textbf{Baoli Sun}$^{5}$ \quad
\textbf{Takahiro Ogawa}$^{2}$ \quad
\textbf{Miki Haseyama}$^{2}$ \quad
\textbf{Zhihui Wang}$^{5,*}$ \\[0.6em]
$^{1}$University of Science and Technology of China \quad
$^{2}$Hokkaido University \\
$^{3}$The University of Queensland \quad
$^{4}$Shanghai AI Laboratory \\
$^{5}$Dalian University of Technology \\[0.4em]
$^{\dagger}$Equal contribution \quad
$^{*}$Corresponding authors \\
\texttt{guang@lmd.ist.hokudai.ac.jp} \quad
\texttt{zhihuiwang@dlut.edu.cn}
}

\begin{document}

\maketitle

\begin{abstract}
Dataset distillation compresses a large training set into a compact synthetic set while preserving its downstream utility. However, existing methods primarily preserve global image statistics and may overlook the localized evidence essential for fine-grained visual classification (FGVC), such as object parts, subtle textures, and region-specific structures. We formulate fine-grained dataset distillation as budgeted discriminative-evidence preservation and propose Discriminative Evidence Composition (DeCO). DeCO uses attention rollout from a pretrained TransFG teacher to identify informative patches, applies spatial diversification to reduce redundant coverage, and organizes the resulting regions into class-wise evidence banks. Multiple same-class regions are then packed into compact grid-composed images. The teacher is used only for dataset construction, whereas downstream students are trained with standard hard-label supervision without teacher logits. Experiments on CUB-200-2011, FGVC-Aircraft, and Stanford Cars show that DeCO consistently outperforms representative coreset and dataset-distillation baselines under different IPC budgets.
\end{abstract}

\section{Introduction}
\label{sec:intro}

Dataset distillation~\cite{li2020soft,DD,li2022awesome, li2022compressed} aims to compress a large training set into a compact dataset that retains its utility for downstream model training. Existing methods typically optimize synthetic samples by matching gradients, features, or training trajectories~\cite{DC,DSA,MTT,li2024dataset,li2024iadd,DATM}, or construct them using pretrained models and feature statistics~\cite{DM,CAFE,SRe2L,DREAM,DCC}. Although these approaches perform well on conventional image classification, their objectives primarily capture global statistics and may overlook the localized evidence required for fine-grained visual classification (FGVC).

In FGVC, category identity is often determined by subtle and spatially localized cues, including object parts, textures, and region-specific structures. Accordingly, FGVC methods have extensively explored discriminative region localization and part-level representation learning~\cite{BCNN,Cross-X,PMG,API-Net,WSDAN2019,CAL,TransFG,DP-Net,CSC-Net,CIN,HBP,iSQRT-COV}. Recent patch-composition methods improve pixel utilization by packing selected local regions into distilled images~\cite{RDED}, while dedicated fine-grained distillation methods introduce localized supervision into the sample optimization process~\cite{FD2}. Nevertheless, existing composition strategies do not explicitly encourage the coverage of spatially distinct fine-grained evidence. Under small IPC budgets, they may therefore preserve background content or repeatedly select neighboring regions while missing complementary class-specific cues.

We formulate fine-grained dataset distillation as budgeted discriminative-evidence preservation: given a fixed image budget, each distilled image should contain dense and spatially diverse class-specific evidence. Based on this perspective, we propose Discriminative Evidence Composition (DeCO). DeCO uses attention rollout from a pretrained TransFG teacher to score candidate patches, applies spatial diversification to reduce redundant coverage, and organizes the selected regions into class-wise evidence banks. Multiple same-class regions are then packed into compact grid-composed images. The teacher is used only during dataset construction, while downstream students are trained with standard hard-label supervision without teacher logits.

We evaluate DeCO on CUB-200-2011, FGVC-Aircraft, and Stanford Cars under different IPC budgets. DeCO consistently outperforms representative coreset and dataset-distillation baselines, while a comparison with a random-region variant demonstrates the importance of discriminative region selection. Our contributions are threefold: (1) we formulate fine-grained dataset distillation from the perspective of budgeted discriminative-evidence preservation, emphasizing both evidence density and spatial coverage; (2) we introduce an attention-guided composition framework that combines discriminative patch scoring, spatial diversification, and class-wise evidence aggregation; and (3) we demonstrate that the resulting distilled datasets support effective hard-label student training across three FGVC benchmarks and different IPC budgets.

\section{Method}
\label{sec:method}

\begin{figure}[t]
    \centering
    \includegraphics[width=0.95\linewidth]{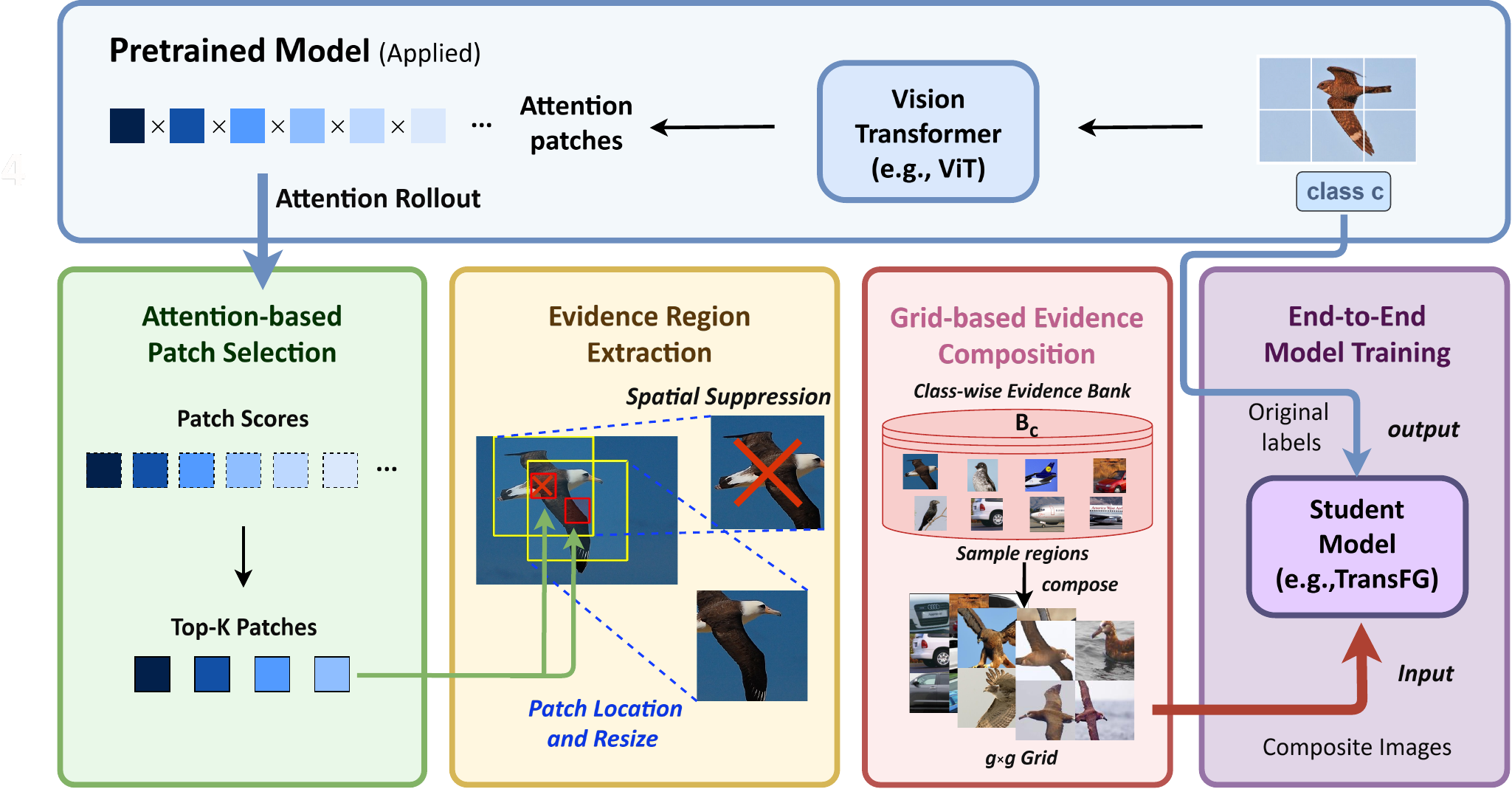}
    \caption{Overview of DeCO. A TransFG teacher scores patches through attention rollout. Spatially diverse high-response regions are collected into class-wise evidence banks and packed into grid-composed images for hard-label student training.}
    \label{fig:deco_pipeline}
\end{figure}

As illustrated in Fig.~\ref{fig:deco_pipeline}, we propose Discriminative Evidence Composition (DeCO), which constructs compact distilled images by preserving localized evidence important for fine-grained recognition. 

\subsection{Discriminative Evidence Mining}
\label{sec:evidence_mining}

Let $\mathcal{T}=\{(x_i,y_i)\}_{i=1}^{|\mathcal{T}|}$ be the original training set with $C$ classes. We train a TransFG teacher on $\mathcal{T}$ and freeze it during dataset construction. Given an image $x$, let $A^{(\ell,h)}$ denote the self-attention matrix of head $h$ at layer $\ell$. Following attention rollout~\cite{abnar2020quantifying}, we first average the attention heads and incorporate the residual connection:
\begin{equation}
\begin{aligned}
\bar{A}^{(\ell)}
    &= \frac{1}{H}\sum_{h=1}^{H}A^{(\ell,h)}, \\
\hat{A}^{(\ell)}
    &= \operatorname{RowNorm}\!\left(\bar{A}^{(\ell)}+I\right), \\
A_{\mathrm{roll}}
    &= \hat{A}^{(L)}\hat{A}^{(L-1)}\cdots\hat{A}^{(1)},
\end{aligned}
\label{eq:attention_rollout}
\end{equation}
where $H$ is the number of attention heads. The evidence score of patch $p$ is then defined by its accumulated attention from the class token:
\begin{equation}
s_p(x)=A_{\mathrm{roll}}[0,p+1].
\label{eq:patch_score}
\end{equation}

Selecting patches solely by $s_p(x)$ may repeatedly cover neighboring regions. We therefore sort the candidates by their scores and apply greedy distance-based spatial suppression. A candidate $p$ is retained only if
\begin{equation}
\left\|(u_p,v_p)-(u_q,v_q)\right\|_2
\geq d_{\min},
\qquad
\forall q\in\mathcal{P}(x),
\label{eq:spatial_suppression}
\end{equation}
where $(u_p,v_p)$ is its spatial center and $\mathcal{P}(x)$ is the set of previously selected patches. This constraint reduces redundant coverage and promotes complementary local evidence.

\subsection{Evidence Bank and Grid Composition}
\label{sec:evidence_composition}

Each selected location is converted into a fixed-size crop $r_p(x_i)$. For class $c$, the corresponding evidence bank is
\begin{equation}
\mathcal{B}_c
=
\left\{
r_p(x_i)
\mid
y_i=c,\;
p\in\mathcal{P}(x_i)
\right\}.
\label{eq:evidence_bank}
\end{equation}
To construct the $j$-th distilled image of class $c$, DeCO samples $M=g^2$ regions from $\mathcal{B}_c$ and arranges them into a $g\times g$ grid:
\begin{equation}
R_{c,j}=\{r_1,\ldots,r_M\}\subseteq\mathcal{B}_c,
\qquad
\tilde{x}_{c,j}=\mathcal{A}(R_{c,j};g),
\label{eq:grid_composition}
\end{equation}
where $\mathcal{A}$ denotes the grid-composition operator. Since all regions are drawn from the same class bank, $\tilde{x}_{c,j}$ is assigned label $c$. This design allocates the fixed pixel budget to multiple discriminative regions rather than a complete image dominated by background. It is particularly useful at IPC$=1$, where a single distilled image must represent an entire class.

\subsection{Hard-label Student Training}
\label{sec:student_training}

After dataset construction, the teacher is discarded and the distilled set
\begin{equation}
\mathcal{S}
=
\left\{
(\tilde{x}_{c,j},c)
\mid
c\in\{1,\ldots,C\},
\;
j\in\{1,\ldots,\mathrm{IPC}\}
\right\}
\label{eq:distilled_set}
\end{equation}
is used to train the downstream student with standard hard-label supervision. No teacher logits, soft labels, or auxiliary distillation losses are required during student training; the teacher is involved only in constructing $\mathcal{S}$. Same-class region composition preserves class-consistent evidence, while spatial suppression reduces redundancy and promotes complementary cues. DeCO can therefore train students with standard hard labels without transferring teacher logits.

\section{Experiments}
\label{sec:experiments}

\subsection{Experimental Settings}

We evaluate DeCO on CUB-200-2011~\cite{CUB-200-2011}, FGVC-Aircraft~\cite{FGVC-Aircraft}, and Stanford Cars~\cite{StanfordCars} under IPC budgets of $\{1,3,5\}$. All images are resized to $224\times224$. A TransFG ViT-B/16 trained on the original training set serves as the teacher for region extraction. A separate TransFG student is randomly initialized and trained from scratch using only the distilled images and standard hard labels. We compare DeCO with Uniform, RDED~\cite{RDED}, SRe$^2$L++~\cite{CVDD}~\cite{SRe2L}, and FADRM+~\cite{FADRM}.

DeCO uses a grid size of $g=2$, corresponding to four regions per distilled image, and a region area ratio of 28\%. The distilled images are generated once and remain fixed during student training. Unless marked otherwise, results are averaged over at least three independent runs.

\begin{table}[t]
\centering
\caption{Top-1 accuracy (\%). Unmarked results are means of at least three runs under a unified TransFG hard-label protocol; $\dagger$ denotes reported results under their original protocols.}
\label{tab:comparison}
\small
\setlength{\tabcolsep}{4.5pt}
\renewcommand{\arraystretch}{1.05}
\begin{tabular}{c c c c c c c c}
\toprule
Dataset & IPC & Uniform & RDED &
SRe$^2$L++$^\dagger$ & FADRM+$^\dagger$ &
DeCO$_{\mathrm{rand}}$ & DeCO \\
\midrule

\multirow{3}{*}{CUB-200-2011}
& 1 & 1.45 & 38.25 & 53.44 & 54.80 & 45.31 & \textbf{65.53} \\
& 3 & 1.97 & 52.56 & 60.04 & 64.06 & 74.60 & \textbf{86.83} \\
& 5 & 2.55 & 63.85 & 63.50 & 66.40 & 78.98 & \textbf{87.99} \\
\cmidrule(lr){2-8}

\multirow{3}{*}{FGVC-Aircraft}
& 1 & 1.82 & 22.11 & 52.60 & 55.02 & 56.02 & \textbf{66.04} \\
& 3 & 3.80 & 36.39 & 66.63 & 72.90 & 76.17 & \textbf{87.49} \\
& 5 & 4.35 & 38.56 & 68.35 & 74.01 & 78.62 & \textbf{88.09} \\
\cmidrule(lr){2-8}

\multirow{3}{*}{Stanford Cars}
& 1 & 1.67 & 16.95 & 52.42 & 60.30 & 46.83 & \textbf{63.94} \\
& 3 & 2.97 & 25.88 & 68.21 & 75.09 & 75.20 & \textbf{85.49} \\
& 5 & 3.18 & 31.86 & 70.90 & 77.71 & 78.30 & \textbf{87.36} \\
\bottomrule
\end{tabular}
\end{table}

\subsection{Main Results}

As shown in Table~\ref{tab:comparison}, DeCO achieves the highest accuracy on all three datasets under every evaluated IPC budget. At IPC$=1$, DeCO reaches 65.53\%, 66.04\%, and 63.94\% on CUB-200-2011, FGVC-Aircraft, and Stanford Cars, respectively. Under the unified evaluation protocol, these results outperform RDED by 27.28, 43.93, and 46.99 percentage points. DeCO also exceeds the strongest reported baseline by 10.73, 11.02, and 3.64 points, although these results were obtained under their original evaluation protocols. The improvements remain consistent as the IPC budget increases, demonstrating the effectiveness of allocating the fixed pixel budget to localized class-specific evidence. DeCO consistently outperforms DeCO$_{\mathrm{rand}}$, which differs only in its region-selection strategy. Since both variants use the same composition and student-training pipeline, the comparison shows that grid composition alone is insufficient and that selecting class-relevant local evidence is critical to DeCO.

\begin{figure}[t]
    \centering
    \includegraphics[width=0.99\linewidth]{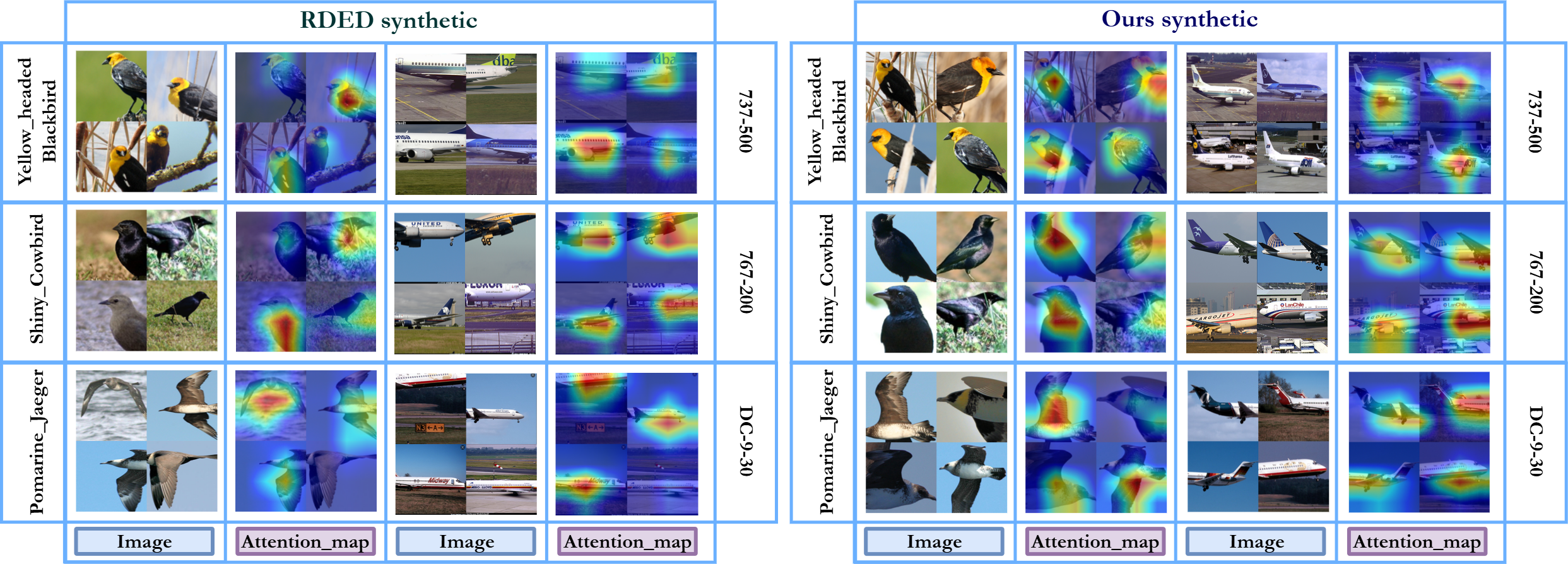}
    \caption{\textbf{Student attention visualization.}
    Grad-CAM~\cite{GradCAM2017} maps of students trained on the RDED and DeCO distilled sets for three CUB-200-2011 categories. Each pair shows an input image and its activation map; DeCO-trained students generally exhibit more localized responses on discriminative object regions.}
    \label{fig:attention_comparison}
\end{figure}

\subsection{Student Attention Visualization}
\label{sec:attention_visualization}

To examine whether the selected local evidence remains useful during downstream training, we apply Grad-CAM~\cite{GradCAM2017} to students trained directly on the distilled datasets. As shown in Fig.~\ref{fig:attention_comparison}, students trained on DeCO generally produce responses that are more concentrated on discriminative object regions than those trained on RDED. In particular, their activations tend to focus on localized object parts rather than being diffusely distributed across the image. These results suggest that the teacher-selected regions remain relevant after dataset construction, even though neither the teacher nor its predictions are used during student training.

\section{Conclusion}
\label{sec:conclusion}

We presented DeCO, a dataset-distillation framework that preserves localized discriminative evidence for fine-grained visual classification. DeCO combines attention-guided region selection, spatial diversification, class-wise evidence aggregation, and grid composition to construct compact distilled images for hard-label student training. Experiments on three fine-grained benchmarks demonstrate consistent improvements under different IPC budgets, while the student visualizations further indicate that the selected local evidence remains relevant during downstream training.

\bibliographystyle{plain}
\bibliography{main}

\clearpage
\appendix

\section{Related Work}
\label{app:related_work}

\paragraph{Dataset Distillation.}
Dataset distillation (DD) aims to construct a compact training set whose downstream utility approaches that of the original data while substantially reducing storage and training costs. Existing methods can be broadly categorized into gradient matching~\cite{DC,DSA,DCC}, distribution matching~\cite{DM,CAFE,li2025davdd}, trajectory matching~\cite{MTT,TESLA,DATM}, decoupled distillation~\cite{SRe2L,FADRM,GVBSM,EDC,LPLD}, and generative distillation~\cite{ITGAN,HPD,Minimax,D4M,wu2025dc3,ye2025igds,li2025diffusion,zou2025dataset,cai2026evlf}. In particular, decoupled methods separate model pretraining, sample construction, and downstream evaluation to improve scalability. SRe$^2$L~\cite{SRe2L} establishes this paradigm through model squeezing, image recovery, and soft relabeling, while SRe$^2$L++~\cite{CVDD} strengthens it with real-image initialization, data augmentation, and batch-specific soft labels. Subsequent methods improve image construction through residual matching~\cite{FADRM}, multi-model statistics~\cite{GVBSM}, or lightweight label representations~\cite{LPLD}. These methods achieve strong performance on general-purpose benchmarks but primarily preserve global statistics and are not specifically designed to retain localized fine-grained evidence.

\paragraph{Patch-based and Fine-grained Dataset Distillation.}
Patch-based methods improve pixel utilization by composing multiple local regions into each distilled image. RDED~\cite{RDED} selects high-confidence crops using an observer model and concatenates them to improve the realism and diversity of distilled data. However, confidence-based selection does not explicitly encourage coverage of spatially distinct regions and may repeatedly retain similar object parts or background content. Recent work has also begun to study DD specifically for fine-grained recognition. FD$^2$~\cite{FD2} incorporates attention-guided fine-grained representations, class prototypes, and diversity constraints into decoupled distillation, improving inter-class separability and within-class diversity. In contrast, DeCO focuses on the construction of hard-label-composed images. It uses class-token attention to identify discriminative regions and spatial suppression to reduce redundant coverage before combining same-class regions under a fixed four-region budget.

\paragraph{Fine-grained Visual Classification.}
Fine-grained visual classification relies on subtle differences between visually similar categories and therefore benefits from localized object parts, textures, and region-specific structures. Cross-X~\cite{Cross-X} models cross-layer and cross-category interactions, while PMG~\cite{PMG} progressively learns multi-granularity representations. DP-Net~\cite{DP-Net} introduces dynamically aligned positional cues, and CSC-Net~\cite{CSC-Net} promotes category-specific semantic coherence. CAL~\cite{CAL} uses counterfactual attention to improve the localization of discriminative regions, whereas TransFG~\cite{TransFG} selects informative patch tokens through transformer attention. Motivated by these observations, DeCO uses a pretrained TransFG teacher as a construction-time evidence localizer. The teacher is discarded after region selection, and the resulting distilled images are used to train randomly initialized students with standard hard-label supervision.

\section{An Evidence-Preservation Analysis}
\label{app:theoretical_perspective}

We provide an informal analysis of why DeCO supports hard-label training on grid-composed images. This analysis is not a formal guarantee of downstream accuracy; rather, it clarifies how evidence strength, spatial diversity, and grid composition jointly affect the reliability of the distilled supervision.

\paragraph{Evidence preservation.}
Let $e(r,c)$ denote the discriminative evidence that region $r$ provides for class $c$. We assume that regions selected from the class-wise evidence bank $\mathcal{B}_c$ contain positive class evidence in expectation:
\begin{equation}
\mu_c
=
\mathbb{E}_{r\sim\mathcal{B}_c}[e(r,c)]
>
0.
\label{eq:positive_evidence}
\end{equation}
For a composed image $\tilde{x}_{c,j}=\mathcal{A}(R_{c,j};g)$, define its average regional evidence as
\begin{equation}
\bar{e}_{c,j}
=
\frac{1}{M}
\sum_{r\in R_{c,j}}e(r,c).
\label{eq:average_evidence}
\end{equation}
We assume that the grid-composition operator approximately preserves this evidence:
\begin{equation}
e(\tilde{x}_{c,j},c)
\geq
\bar{e}_{c,j}
-
\epsilon_{\mathcal{A}},
\label{eq:evidence_preserving}
\end{equation}
where $\epsilon_{\mathcal{A}}\geq0$ represents the distortion introduced by cropping, resizing, and grid composition. Taking expectations gives
\begin{equation}
\mathbb{E}[e(\tilde{x}_{c,j},c)]
\geq
\mu_c-\epsilon_{\mathcal{A}}.
\label{eq:expected_composed_evidence}
\end{equation}
Thus, a composed image retains positive expected class evidence whenever $\mu_c>\epsilon_{\mathcal{A}}$, providing an intuitive justification for assigning it the original hard label $c$.

\paragraph{Evidence concentration under spatial diversity.}
Positive expected evidence does not by itself guarantee that every composed image is informative. Let $e_m=e(r_m,c)$ be the evidence provided by the $m$-th selected region. We assume
\begin{equation}
\mathrm{Var}[e_m]\leq\sigma_c^2,
\qquad
\mathrm{Cov}[e_m,e_{m'}]
\leq
\rho_c\sigma_c^2,
\quad m\neq m',
\label{eq:evidence_assumptions}
\end{equation}
where $\rho_c\in[0,1]$ controls the redundancy between selected regions. The variance of their empirical average satisfies
\begin{align}
\mathrm{Var}[\bar{e}_{c,j}]
&=
\frac{1}{M^2}
\left(
\sum_{m=1}^{M}\mathrm{Var}[e_m]
+
2\sum_{m<m'}\mathrm{Cov}[e_m,e_{m'}]
\right)
\nonumber\\
&\leq
\frac{\sigma_c^2}{M}
\left[1+(M-1)\rho_c\right].
\label{eq:evidence_variance_bound}
\end{align}

\noindent\textbf{Proposition 1 (Evidence concentration).}
Suppose $\mu_c>\epsilon_{\mathcal{A}}$ and the conditions in Eq.~\eqref{eq:evidence_assumptions} hold. The probability that a composed image fails to retain positive class evidence is bounded by
\begin{equation}
\Pr\!\left[e(\tilde{x}_{c,j},c)\leq0\right]
\leq
\frac{
\sigma_c^2\left[1+(M-1)\rho_c\right]
}{
M(\mu_c-\epsilon_{\mathcal{A}})^2
}.
\label{eq:evidence_failure_bound}
\end{equation}

\noindent\emph{Proof.}
From Eq.~\eqref{eq:evidence_preserving}, the event
$e(\tilde{x}_{c,j},c)\leq0$ implies
$\bar{e}_{c,j}\leq\epsilon_{\mathcal{A}}$. Chebyshev's inequality therefore gives
\begin{align}
\Pr[e(\tilde{x}_{c,j},c)\leq0]
&\leq
\Pr[\bar{e}_{c,j}\leq\epsilon_{\mathcal{A}}]
\nonumber\\
&\leq
\frac{\mathrm{Var}[\bar{e}_{c,j}]}
{(\mu_c-\epsilon_{\mathcal{A}})^2}.
\end{align}
Substituting Eq.~\eqref{eq:evidence_variance_bound} yields Eq.~\eqref{eq:evidence_failure_bound}.
\hfill$\square$

\paragraph{Interpretation.}
Equation~\ref{eq:evidence_failure_bound} reveals three complementary effects. First, attention-guided selection is intended to increase $\mu_c$ by favoring regions with stronger class evidence. Second, spatial suppression discourages neighboring crops and is expected to reduce $\rho_c$, thereby limiting evidence redundancy. Third, composing multiple regions increases $M$ and improves evidence concentration when their correlations are controlled.

This relationship can be expressed through the effective number of evidence regions:
\begin{equation}
M_{\mathrm{eff}}
=
\frac{M}{1+(M-1)\rho_c}.
\label{eq:effective_regions}
\end{equation}
The failure bound decreases proportionally to $1/M_{\mathrm{eff}}$. Therefore, increasing the number of composed regions provides limited benefit when their evidence is highly correlated. In our implementation, DeCO uses $M=4$ regions per image, while spatial suppression encourages these regions to contain complementary rather than repetitive evidence.

The distortion term $\epsilon_{\mathcal{A}}$ also clarifies the role of region size. Very small crops may reduce $\mu_c$ by removing structural context, whereas overly large crops introduce background and redundant content. An intermediate crop size therefore balances evidence strength and composition distortion, consistent with the empirical results in Appendix~\ref{app:region_ratio}.

\section{Additional Experimental Results}
\label{app:additional_exp}

We analyze the sensitivity of DeCO to the region area ratio, teacher patch size, and attention rollout depth, followed by a qualitative comparison with RDED. Unless otherwise stated, all experiments use IPC$=1$, a randomly initialized TransFG student, and the same hard-label training protocol as the main experiments. Reported quantitative results are averaged over at least three independent runs.

\subsection{Impact of Region Area Ratio}
\label{app:region_ratio}

The region area ratio controls the trade-off between preserving localized discriminative cues and retaining sufficient contextual information. Very small regions preserve object parts or local textures but may discard important structural information. Conversely, excessively large regions introduce more background content and reduce the density of discriminative evidence.

\begin{figure}[t]
\centering
\includegraphics[width=0.5\linewidth]{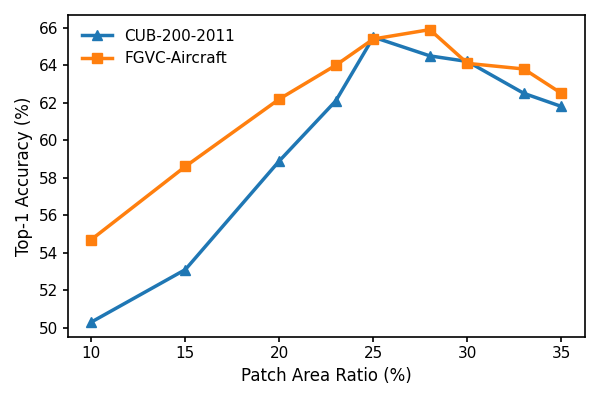}
\caption{Mean top-1 accuracy under different cropped-region area ratios on CUB-200-2011 and FGVC-Aircraft.}
\label{fig:patch_ablation}
\end{figure}

As shown in Fig.~\ref{fig:patch_ablation}, intermediate region sizes provide a favorable balance between local evidence and contextual information. Although the optimal ratio varies slightly across datasets, performance remains stable within the intermediate range. We use a common region area ratio of 28\% for all benchmarks in the main experiments.

\subsection{Impact of Initial Patch Size}
\label{app:patch_size}

We study whether DeCO is sensitive to the initial patch granularity of the TransFG teacher by comparing ViT-B/16 and ViT-B/32, whose patch sizes are $16\times16$ and $32\times32$, respectively.

\begin{wraptable}{r}{0.42\linewidth}
\vspace{-12pt}
\centering
\caption{Mean top-1 accuracy (\%) under different teacher patch sizes.}
\label{tab:patch_size}
\vspace{-6pt}
\scriptsize
\resizebox{\linewidth}{!}{
\begin{tabular}{lcc}
\toprule
Dataset & ViT-B/16 & ViT-B/32 \\
\midrule
CUB-200-2011 & 65.4 & 65.2 \\
FGVC-Aircraft & 66.0 & 66.1 \\
\bottomrule
\end{tabular}
}
\vspace{-12pt}
\end{wraptable}

As shown in Table~\ref{tab:patch_size}, changing the teacher patch size alters accuracy by at most 0.2 percentage points, indicating limited sensitivity to the initial patch resolution. The selected patch primarily determines the center of a larger evidence crop; moderate changes in localization granularity therefore produce similar regions after cropping and resizing.

\subsection{Sensitivity to Attention Rollout Depth}
\label{app:rollout_depth}

We next vary the number of Transformer blocks included in attention rollout. Consistent with Eq.~\eqref{eq:attention_rollout}, cumulative attention up to block $l$ is computed as
\begin{equation}
A_{\mathrm{roll}}^{(l)}
=
\hat{A}^{(l)}
\hat{A}^{(l-1)}
\cdots
\hat{A}^{(1)},
\qquad
l\in\{1,3,5,7,9,11\}.
\label{eq:rollout_depth}
\end{equation}
Here, $\hat{A}^{(l)}$ denotes the head-averaged and residual-normalized attention matrix at block $l$. Only the rollout depth is varied; region extraction, evidence-bank construction, composition, and student training remain unchanged. We evaluate up to the eleventh block because TransFG uses the first 11 blocks for part selection before processing the selected tokens with the final block.

\begin{table}[t]
\centering
\caption{Sensitivity to attention rollout depth. Results are mean top-1 accuracy (\%) at IPC$=1$.}
\label{tab:rollout_depth}
\small
\setlength{\tabcolsep}{20pt}
\renewcommand{\arraystretch}{1.15}
\begin{tabular}{c c c}
\toprule
Rollout Depth $l$ & CUB-200-2011 & FGVC-Aircraft \\
\midrule
1  & 64.12 & 65.18 \\
3  & 64.93 & 65.71 \\
5  & 65.38 & 65.96 \\
7  & 65.47 & 66.02 \\
9  & 65.51 & 66.01 \\
11 & \textbf{65.53} & \textbf{66.04} \\
\bottomrule
\end{tabular}
\end{table}

Table~\ref{tab:rollout_depth} shows that performance improves as attention is accumulated through the early and middle blocks and becomes stable at later depths. On CUB-200-2011, accuracy changes by only 0.15 percentage points between depths 5 and 11; the corresponding difference on FGVC-Aircraft is 0.08 points. We therefore use rollout through the eleventh block as the default configuration.

\subsection{Qualitative Comparison with RDED}
\label{app:qualitative_rded}

\begin{figure}[t]
    \centering
    \includegraphics[width=0.93\linewidth]{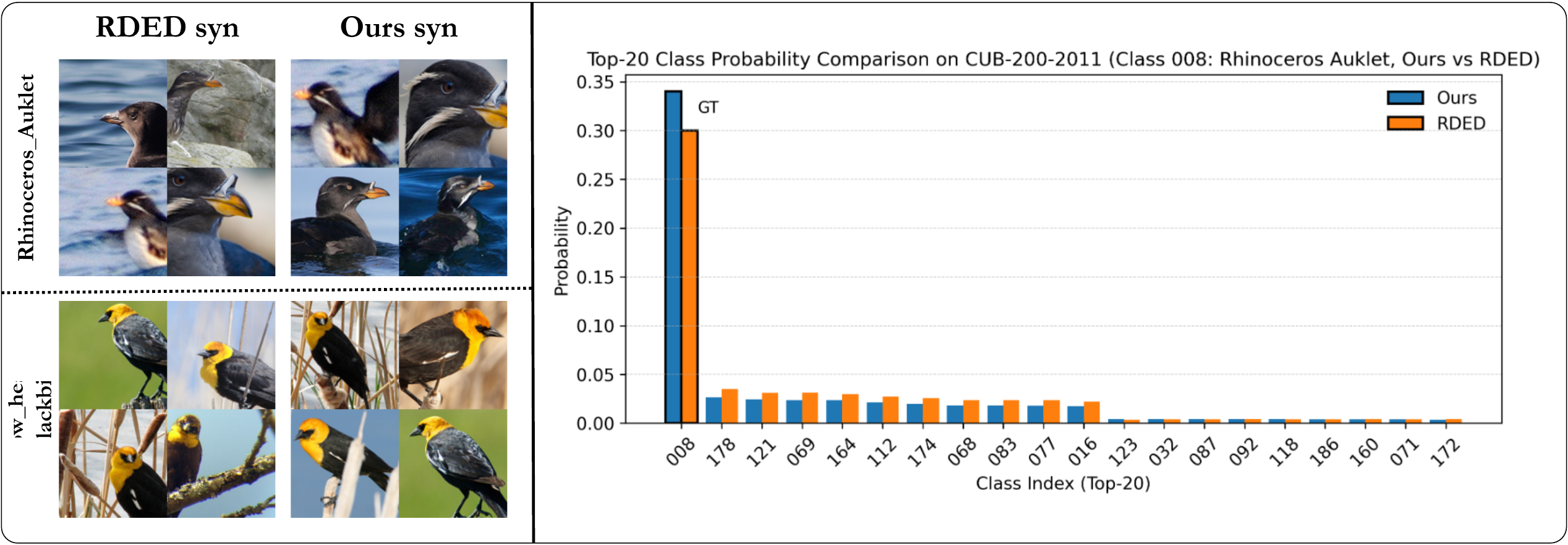}
    \caption{\textbf{Qualitative comparison with RDED.}
    Left: distilled images produced by RDED and DeCO using the same four-region budget. DeCO retains more localized class-specific evidence, whereas RDED may preserve larger background or weakly localized regions. Right: top-20 predicted class probabilities for a representative CUB-200-2011 category.}
    \label{fig:rdeco_comparison}
\end{figure}

Figure~\ref{fig:rdeco_comparison} compares RDED and DeCO under the same four-region composition budget. RDED can retain relatively large background regions or weakly localized content, whereas DeCO uses teacher attention and spatial suppression to select complementary class-specific regions. The comparison illustrates that DeCO changes the information retained within the fixed pixel budget rather than increasing the number of composed regions.

\section{Visualization of Distilled Samples}
\label{sec:visualization}

\begin{figure}[p]
    \centering
    \includegraphics[
        width=\linewidth,
        height=0.82\textheight,
        keepaspectratio
    ]{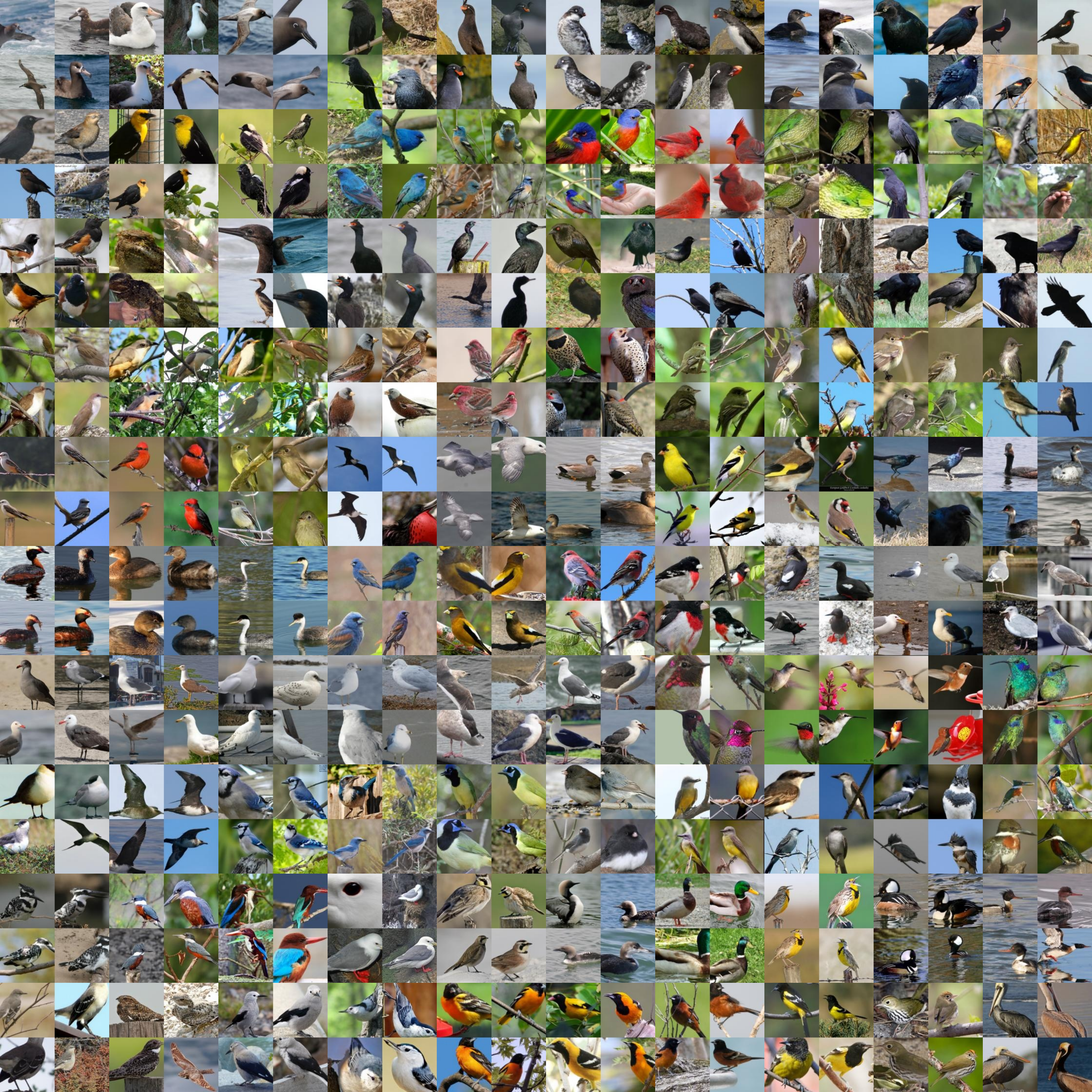}
    \caption{Visualization of distilled samples from the first 100 classes of CUB-200-2011 at IPC=1.}
    \label{fig:cub_first100}
\end{figure}

\begin{figure}[p]
    \centering
    \includegraphics[
        width=\linewidth,
        height=0.82\textheight,
        keepaspectratio
    ]{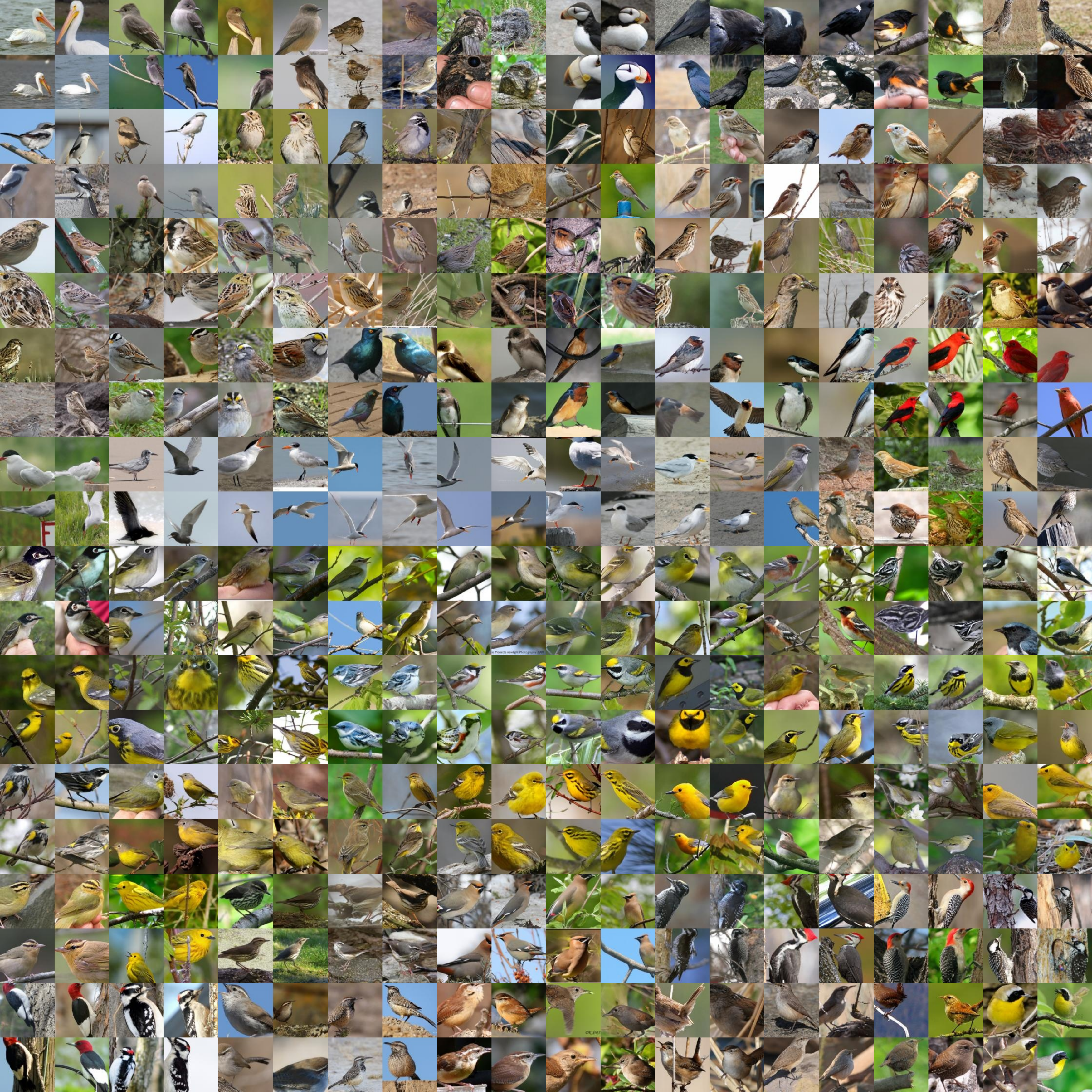}
    \caption{Visualization of distilled samples from the last 100 classes of CUB-200-2011 at IPC=1.}
    \label{fig:cub_last100}
\end{figure}

\begin{figure}[p]
    \centering
    \includegraphics[
        width=\linewidth,
        height=0.82\textheight,
        keepaspectratio
    ]{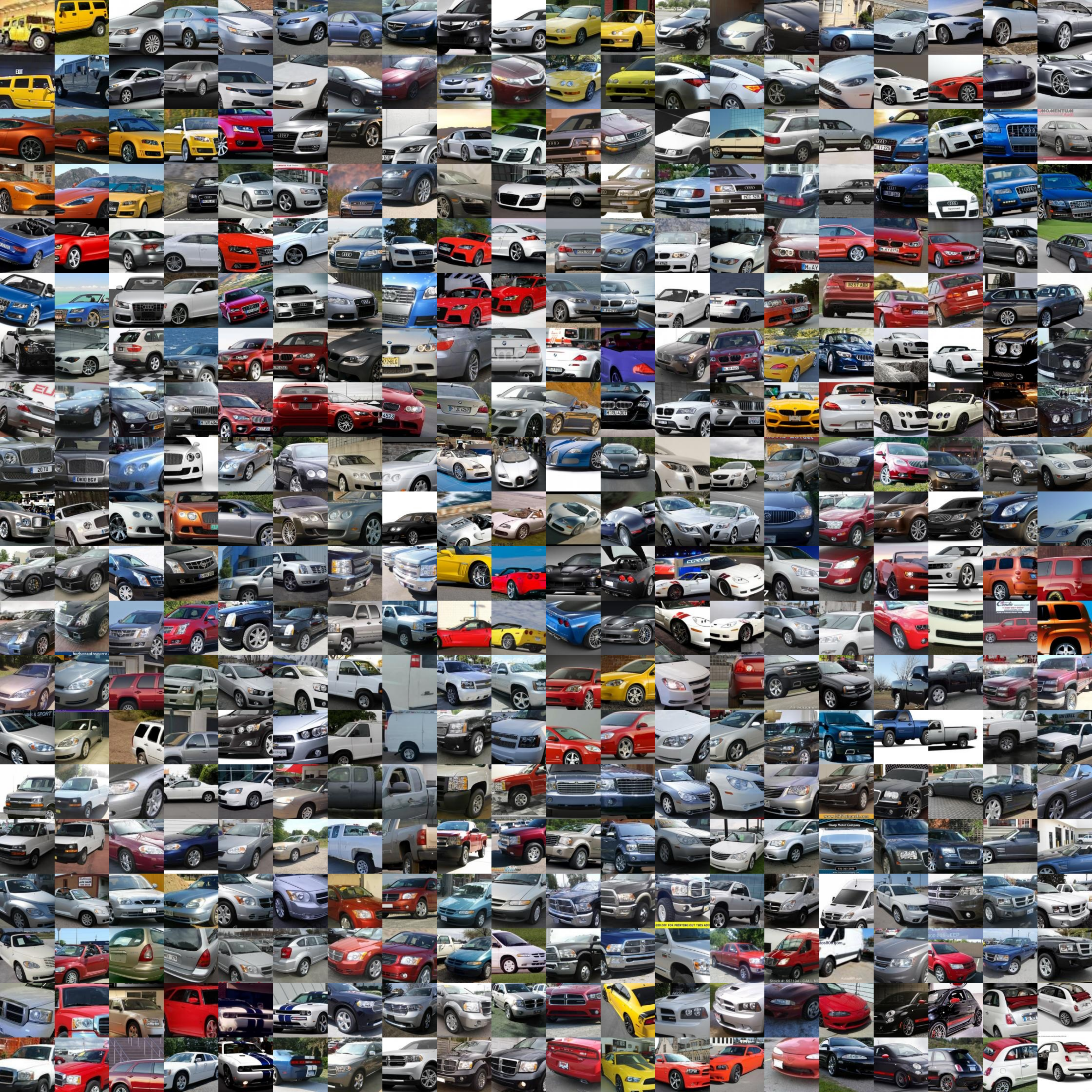}
    \caption{Visualization of distilled samples from the first 100 classes of Stanford Cars at IPC=1.}
    \label{fig:cars_first100}
\end{figure}

\begin{figure}[p]
    \centering
    \includegraphics[
        width=\linewidth,
        height=0.82\textheight,
        keepaspectratio
    ]{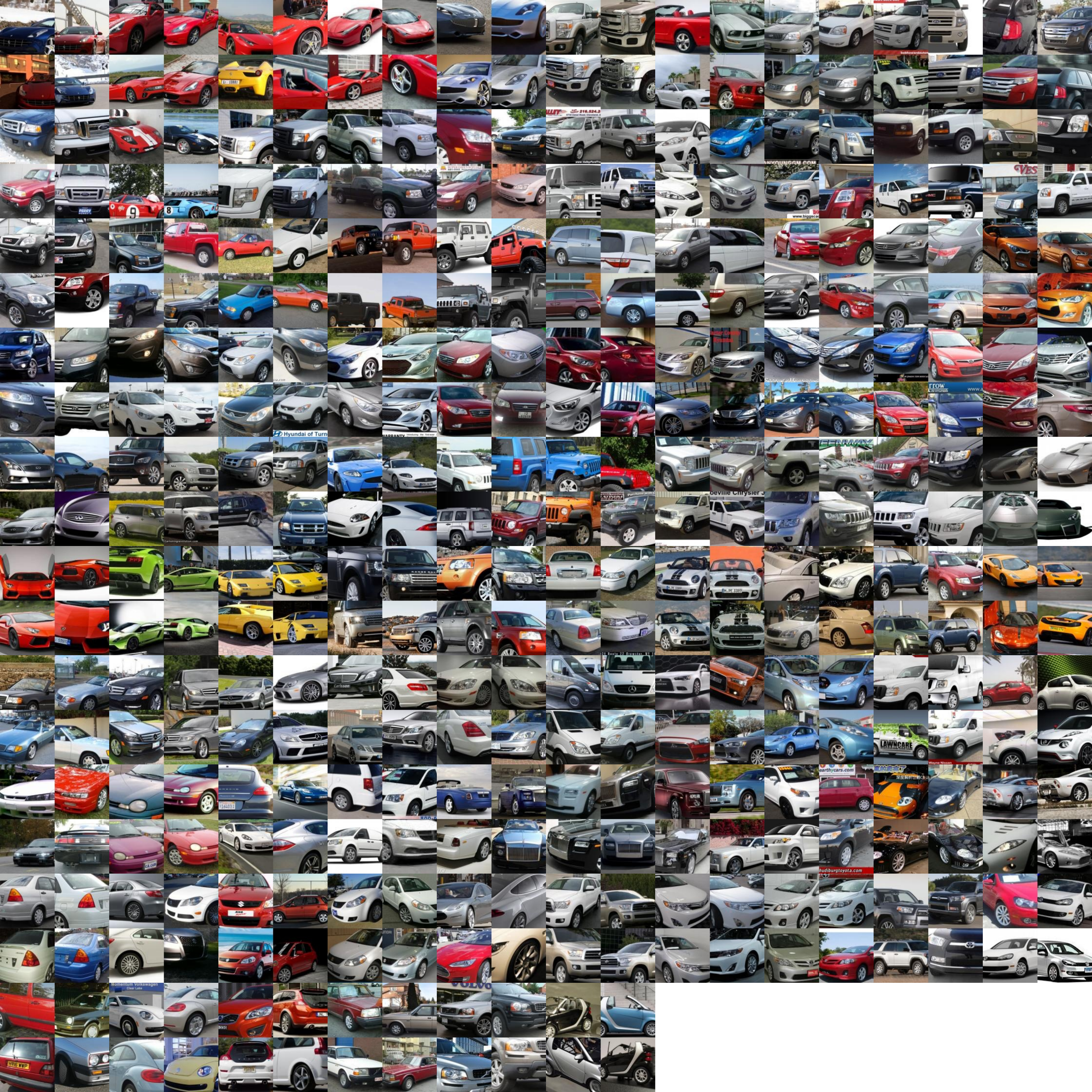}
    \caption{Visualization of distilled samples from the remaining 96 classes of Stanford Cars at IPC=1.}
    \label{fig:cars_last96}
\end{figure}

\begin{figure}[p]
    \centering
    \includegraphics[
        width=\linewidth,
        height=0.82\textheight,
        keepaspectratio
    ]{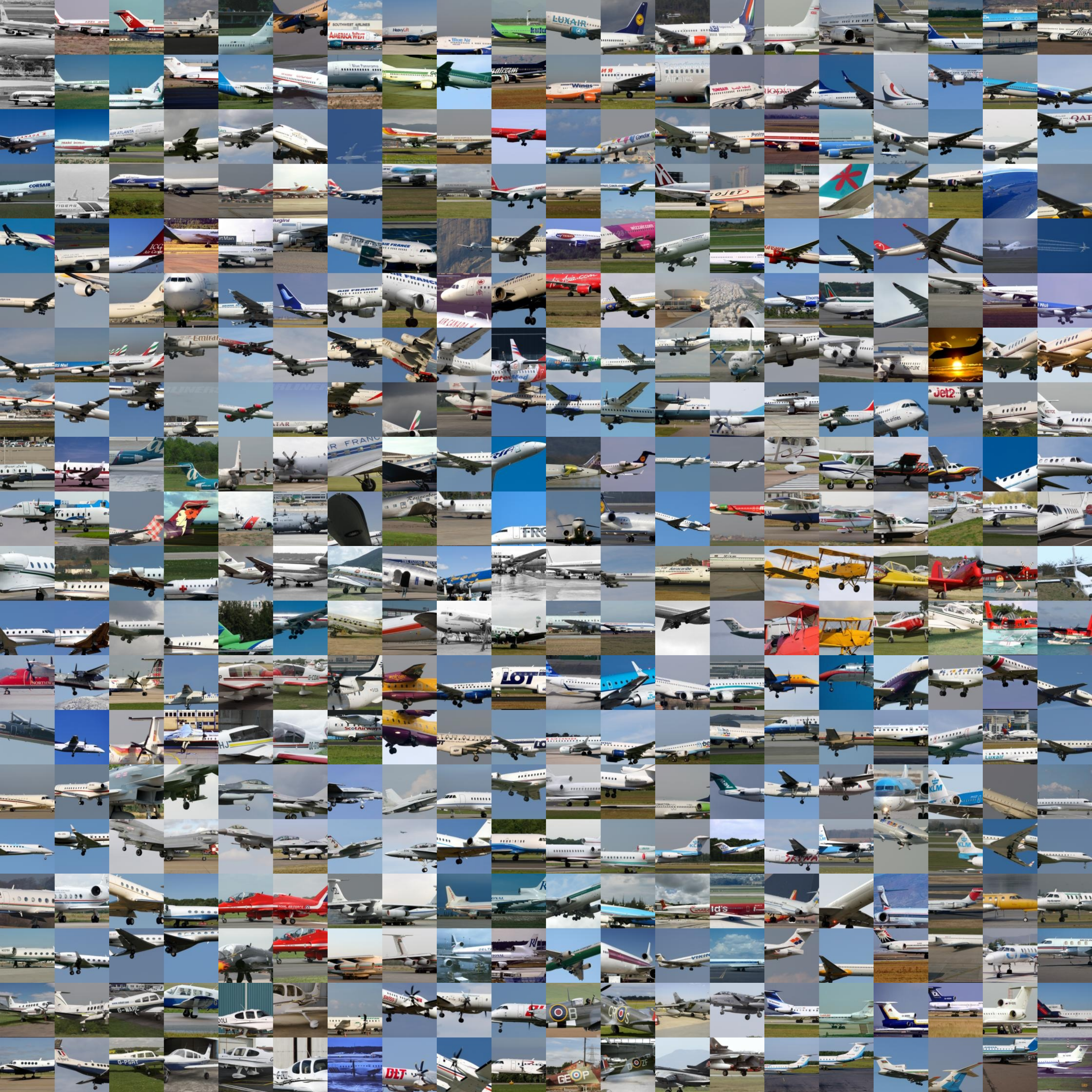}
    \caption{Visualization of distilled samples from all 100 classes of FGVC-Aircraft at IPC=1.}
    \label{fig:aircraft_all}
\end{figure}

At IPC=1, the fine-grained samples distilled by DeCO are shown in Figs.~\ref{fig:cub_first100} and \ref{fig:cub_last100} for CUB-200-2011, Figs.~\ref{fig:cars_first100} and \ref{fig:cars_last96} for Stanford Cars, and Fig.~\ref{fig:aircraft_all} for FGVC-Aircraft.

\clearpage

\clearpage

\clearpage

\clearpage

\clearpage

\newpage
\section*{NeurIPS Paper Checklist}

\begin{enumerate}

\item {\bf Claims}
    \item[] Question: Do the main claims made in the abstract and introduction accurately reflect the paper's contributions and scope?
    \item[] Answer: \answerYes{}
    \item[] Justification: The abstract and introduction accurately describe the proposed DeCO framework, its focus on preserving discriminative evidence for fine-grained dataset distillation, and its evaluation under low-data budgets. The claims are supported by the experimental results on CUB-200-2011, FGVC-Aircraft, and Stanford Cars, without asserting broader generalization beyond the evaluated settings.

    \item[] Guidelines:
    \begin{itemize}
        \item The answer \answerNA{} means that the abstract and introduction do not include the claims made in the paper.
        \item The abstract and/or introduction should clearly state the claims made, including the contributions made in the paper and important assumptions and limitations. A \answerNo{} or \answerNA{} answer to this question will not be perceived well by the reviewers.
        \item The claims made should match theoretical and experimental results, and reflect how much the results can be expected to generalize to other settings.
        \item It is fine to include aspirational goals as motivation as long as it is clear that these goals are not attained by the paper.
    \end{itemize}

\item {\bf Limitations}
    \item[] Question: Does the paper discuss the limitations of the work performed by the authors?
    \item[] Answer: \answerNo{}
    \item[] Justification: The paper does not include a dedicated discussion of limitations. The proposed method is evaluated on three fine-grained visual classification benchmarks and under limited IPC budgets, and its dependence on a pretrained teacher and construction-stage computational cost are not explicitly discussed in the current manuscript.

    \item[] Guidelines:
    \begin{itemize}
        \item The answer \answerNA{} means that the paper has no limitation while the answer \answerNo{} means that the paper has limitations, but those are not discussed in the paper.
        \item The authors are encouraged to create a separate ``Limitations'' section in their paper.
        \item The paper should point out any strong assumptions and how robust the results are to violations of these assumptions (e.g., independence assumptions, noiseless settings, model well-specification, asymptotic approximations only holding locally). The authors should reflect on how these assumptions might be violated in practice and what the implications would be.
        \item The authors should reflect on the scope of the claims made, e.g., if the approach was only tested on a few datasets or with a few runs. In general, empirical results often depend on implicit assumptions, which should be articulated.
        \item The authors should reflect on the factors that influence the performance of the approach. For example, a facial recognition algorithm may perform poorly when image resolution is low or images are taken in low lighting. Or a speech-to-text system might not be used reliably to provide closed captions for online lectures because it fails to handle technical jargon.
        \item The authors should discuss the computational efficiency of the proposed algorithms and how they scale with dataset size.
        \item If applicable, the authors should discuss possible limitations of their approach to address problems of privacy and fairness.
        \item While the authors might fear that complete honesty about limitations might be used by reviewers as grounds for rejection, a worse outcome might be that reviewers discover limitations that aren't acknowledged in the paper. The authors should use their best judgment and recognize that individual actions in favor of transparency play an important role in developing norms that preserve the integrity of the community. Reviewers will be specifically instructed to not penalize honesty concerning limitations.
    \end{itemize}

\item {\bf Theory assumptions and proofs}
    \item[] Question: For each theoretical result, does the paper provide the full set of assumptions and a complete (and correct) proof?
    \item[] Answer: \answerYes{}
    \item[] Justification: Appendix~\ref{app:theoretical_perspective} explicitly states the assumptions used by Proposition~1 and provides a complete proof of the resulting evidence-failure bound using Chebyshev's inequality. The assumptions are stated in Eq.~\eqref{eq:evidence_assumptions}, and the proposition and proof are cross-referenced to the relevant equations.

    \item[] Guidelines:
    \begin{itemize}
        \item The answer \answerNA{} means that the paper does not include theoretical results.
        \item All the theorems, formulas, and proofs in the paper should be numbered and cross-referenced.
        \item All assumptions should be clearly stated or referenced in the statement of any theorems.
        \item The proofs can either appear in the main paper or the supplemental material, but if they appear in the supplemental material, the authors are encouraged to provide a short proof sketch to provide intuition.
        \item Inversely, any informal proof provided in the core of the paper should be complemented by formal proofs provided in appendix or supplemental material.
        \item Theorems and Lemmas that the proof relies upon should be properly referenced.
    \end{itemize}

\item {\bf Experimental result reproducibility}
    \item[] Question: Does the paper fully disclose all the information needed to reproduce the main experimental results of the paper to the extent that it affects the main claims and/or conclusions of the paper (regardless of whether the code and data are provided or not)?
    \item[] Answer: \answerYes{}
    \item[] Justification: The paper provides the datasets and evaluation protocol, teacher architecture, evidence extraction procedure, patch selection and composition strategy, IPC settings, student training protocol, and baseline configurations needed to reproduce the main experimental results. Additional implementation details are provided in the appendix and supplementary material.

    \item[] Guidelines:
    \begin{itemize}
        \item The answer \answerNA{} means that the paper does not include experiments.
        \item If the paper includes experiments, a \answerNo{} answer to this question will not be perceived well by the reviewers: Making the paper reproducible is important, regardless of whether the code and data are provided or not.
        \item If the contribution is a dataset and\slash or model, the authors should describe the steps taken to make their results reproducible or verifiable.
        \item Depending on the contribution, reproducibility can be accomplished in various ways. For example, if the contribution is a novel architecture, describing the architecture fully might suffice, or if the contribution is a specific model and empirical evaluation, it may be necessary to either make it possible for others to replicate the model with the same dataset, or provide access to the model. In general. releasing code and data is often one good way to accomplish this, but reproducibility can also be provided via detailed instructions for how to replicate the results, access to a hosted model (e.g., in the case of a large language model), releasing of a model checkpoint, or other means that are appropriate to the research performed.
        \item While NeurIPS does not require releasing code, the conference does require all submissions to provide some reasonable avenue for reproducibility, which may depend on the nature of the contribution. For example
        \begin{enumerate}
            \item If the contribution is primarily a new algorithm, the paper should make it clear how to reproduce that algorithm.
            \item If the contribution is primarily a new model architecture, the paper should describe the architecture clearly and fully.
            \item If the contribution is a new model (e.g., a large language model), then there should either be a way to access this model for reproducing the results or a way to reproduce the model (e.g., with an open-source dataset or instructions for how to construct the dataset).
            \item We recognize that reproducibility may be tricky in some cases, in which case authors are welcome to describe the particular way they provide for reproducibility. In the case of closed-source models, it may be that access to the model is limited in some way (e.g., to registered users), but it should be possible for other researchers to have some path to reproducing or verifying the results.
        \end{enumerate}
    \end{itemize}

\item {\bf Open access to data and code}
    \item[] Question: Does the paper provide open access to the data and code, with sufficient instructions to faithfully reproduce the main experimental results, as described in supplemental material?
    \item[] Answer: \answerNo{}
    \item[] Justification: The experiments use publicly available benchmark datasets, but the authors do not provide an open-access code repository or a public release of the implementation at submission time. The paper nevertheless provides sufficient methodological and experimental details for reproducing the reported results, and the datasets used in the experiments are publicly accessible from their original sources.

    \item[] Guidelines:
    \begin{itemize}
        \item The answer \answerNA{} means that paper does not include experiments requiring code.
        \item Please see the NeurIPS code and data submission guidelines (\url{https://neurips.cc/public/guides/CodeSubmissionPolicy}) for more details.
        \item While we encourage the release of code and data, we understand that this might not be possible, so \answerNo{} is an acceptable answer. Papers cannot be rejected simply for not including code, unless this is central to the contribution (e.g., for a new open-source benchmark).
        \item The instructions should contain the exact command and environment needed to run to reproduce the results. See the NeurIPS code and data submission guidelines (\url{https://neurips.cc/public/guides/CodeSubmissionPolicy}) for more details.
        \item The authors should provide instructions on data access and preparation, including how to access the raw data, preprocessed data, intermediate data, and generated data, etc.
        \item The authors should provide scripts to reproduce all experimental results for the new proposed method and baselines. If only a subset of experiments are reproducible, they should state which ones are omitted from the script and why.
        \item At submission time, to preserve anonymity, the authors should release anonymized versions (if applicable).
        \item Providing as much information as possible in supplemental material (appended to the paper) is recommended, but including URLs to data and code is permitted.
    \end{itemize}

\item {\bf Experimental setting/details}
    \item[] Question: Does the paper specify all the training and test details (e.g., data splits, hyperparameters, how they were chosen, type of optimizer) necessary to understand the results?
    \item[] Answer: \answerYes{}
    \item[] Justification: The paper specifies the datasets and splits, teacher and student architectures, image preprocessing, IPC settings, evidence extraction and composition procedure, optimization settings, and evaluation protocol. Additional implementation and hyperparameter details are provided in the appendix and supplementary material.

    \item[] Guidelines:
    \begin{itemize}
        \item The answer \answerNA{} means that the paper does not include experiments.
        \item The experimental setting should be presented in the core of the paper to a level of detail that is necessary to appreciate the results and make sense of them.
        \item The full details can be provided either with the code, in appendix, or as supplemental material.
    \end{itemize}

\item {\bf Experiment statistical significance}
    \item[] Question: Does the paper report error bars suitably and correctly defined or other appropriate information about the statistical significance of the experiments?
    \item[] Answer: \answerNo{}
    \item[] Justification: The main experimental results are reported as point estimates without error bars, confidence intervals, or formal statistical significance tests. Reporting multiple independent runs for all combinations of datasets, IPC settings, methods, and experimental configurations would substantially increase the computational cost, and the paper does not claim statistical significance from the reported point estimates.

    \item[] Guidelines:
    \begin{itemize}
        \item The answer \answerNA{} means that the paper does not include experiments.
        \item The authors should answer \answerYes{} if the results are accompanied by error bars, confidence intervals, or statistical significance tests, at least for the experiments that support the main claims of the paper.
        \item The factors of variability that the error bars are capturing should be clearly stated (for example, train/test split, initialization, random drawing of some parameter, or overall run with given experimental conditions).
        \item The method for calculating the error bars should be explained (closed form formula, call to a library function, bootstrap, etc.)
        \item The assumptions made should be given (e.g., Normally distributed errors).
        \item It should be clear whether the error bar is the standard deviation or the standard error of the mean.
        \item It is OK to report 1-sigma error bars, but one should state it. The authors should preferably report a 2-sigma error bar than state that they have a 96\% CI, if the hypothesis of Normality of errors is not verified.
        \item For asymmetric distributions, the authors should be careful not to show in tables or figures symmetric error bars that would yield results that are out of range (e.g., negative error rates).
        \item If error bars are reported in tables or plots, the authors should explain in the text how they were calculated and reference the corresponding figures or tables in the text.
    \end{itemize}

\item {\bf Experiments compute resources}
    \item[] Question: For each experiment, does the paper provide sufficient information on the computer resources (type of compute workers, memory, time of execution) needed to reproduce the experiments?
    \item[] Answer: \answerNo{}
    \item[] Justification: The current manuscript does not report the full hardware configuration, per-experiment execution time, or total compute. Therefore, the information is not sufficient to fully satisfy the requested compute-resource disclosure.

    \item[] Guidelines:
    \begin{itemize}
        \item The answer \answerNA{} means that the paper does not include experiments.
        \item The paper should indicate the type of compute workers CPU or GPU, internal cluster, or cloud provider, including relevant memory and storage.
        \item The paper should provide the amount of compute required for each of the individual experimental runs as well as estimate the total compute.
        \item The paper should disclose whether the full research project required more compute than the experiments reported in the paper (e.g., preliminary or failed experiments that didn't make it into the paper).
    \end{itemize}

\item {\bf Code of ethics}
    \item[] Question: Does the research conducted in the paper conform, in every respect, with the NeurIPS Code of Ethics \url{https://neurips.cc/public/EthicsGuidelines}?
    \item[] Answer: \answerYes{}
    \item[] Justification: The work uses publicly available benchmark datasets and pretrained models for research purposes and does not involve human-subject experiments, collection of personal information, or high-risk deployment. The research is conducted in accordance with the principles described in the NeurIPS Code of Ethics.

    \item[] Guidelines:
    \begin{itemize}
        \item The answer \answerNA{} means that the authors have not reviewed the NeurIPS Code of Ethics.
        \item If the authors answer \answerNo, they should explain the special circumstances that require a deviation from the Code of Ethics.
        \item The authors should make sure to preserve anonymity (e.g., if there is a special consideration due to laws or regulations in their jurisdiction).
    \end{itemize}

\item {\bf Broader impacts}
    \item[] Question: Does the paper discuss both potential positive societal impacts and negative societal impacts of the work performed?
    \item[] Answer: \answerNo{}
    \item[] Justification: The work has potential positive impacts by reducing the storage and training cost associated with fine-grained visual recognition, which may improve the accessibility and efficiency of machine learning experiments. However, the paper does not provide a dedicated discussion of both positive and negative societal impacts, and therefore does not fully satisfy this checklist item.

    \item[] Guidelines:
    \begin{itemize}
        \item The answer \answerNA{} means that there is no societal impact of the work performed.
        \item If the authors answer \answerNA{} or \answerNo, they should explain why their work has no societal impact or why the paper does not address societal impact.
        \item Examples of negative societal impacts include potential malicious or unintended uses (e.g., disinformation, generating fake profiles, surveillance), fairness considerations (e.g., deployment of technologies that could make decisions that unfairly impact specific groups), privacy considerations, and security considerations.
        \item The conference expects that many papers will be foundational research and not tied to particular applications, let alone deployments. However, if there is a direct path to any negative applications, the authors should point it out. For example, it is legitimate to point out that an improvement in the quality of generative models could be used to generate Deepfakes for disinformation. On the other hand, it is not needed to point out that a generic algorithm for optimizing neural networks could enable people to train models that generate Deepfakes faster.
        \item The authors should consider possible harms that could arise when the technology is being used as intended and functioning correctly, harms that could arise when the technology is being used as intended but gives incorrect results, and harms following from (intentional or unintentional) misuse of the technology.
        \item If there are negative societal impacts, the authors could also discuss possible mitigation strategies (e.g., gated release of models, providing defenses in addition to attacks, mechanisms for monitoring misuse, mechanisms to monitor how a system learns from feedback over time, improving the efficiency and accessibility of ML).
    \end{itemize}

\item {\bf Safeguards}
    \item[] Question: Does the paper describe safeguards that have been put in place for responsible release of data or models that have a high risk for misuse (e.g., pre-trained language models, image generators, or scraped datasets)?
    \item[] Answer: \answerNA{}
    \item[] Justification: The paper does not release a high-risk pretrained language model, image generator, surveillance system, or other asset with a substantial foreseeable misuse risk requiring the safeguards described in this question. The proposed method is a dataset distillation algorithm evaluated on standard fine-grained visual recognition benchmarks.

    \item[] Guidelines:
    \begin{itemize}
        \item The answer \answerNA{} means that the paper poses no such risks.
        \item Released models that have a high risk for misuse or dual-use should be released with necessary safeguards to allow for controlled use of the model, for example by requiring that users adhere to usage guidelines or restrictions to access the model or implementing safety filters.
        \item Datasets that have been scraped from the Internet could pose safety risks. The authors should describe how they avoided releasing unsafe images.
        \item We recognize that providing effective safeguards is challenging, and many papers do not require this, but we encourage authors to take this into account and make a best faith effort.
    \end{itemize}

\item {\bf Licenses for existing assets}
    \item[] Question: Are the creators or original owners of assets (e.g., code, data, models), used in the paper, properly credited and are the license and terms of use explicitly mentioned and properly respected?
    \item[] Answer: \answerNo{}
    \item[] Justification: The paper cites the original datasets and pretrained models used in the experiments, but the license and terms of use are not comprehensively documented for every existing asset in the current submission. The relevant dataset and model licenses should be explicitly listed in the final version or supplementary material where applicable.

    \item[] Guidelines:
    \begin{itemize}
        \item The answer \answerNA{} means that the paper does not use existing assets.
        \item The authors should cite the original paper that produced the code package or dataset.
        \item The authors should state which version of the asset is used and, if possible, include a URL.
        \item The name of the license (e.g., CC-BY 4.0) should be included for each asset.
        \item For scraped data from a particular source (e.g., website), the copyright and terms of service of that source should be provided.
        \item If assets are released, the license, copyright information, and terms of use in the package should be provided. For popular datasets, \url{paperswithcode.com/datasets} has curated licenses for some datasets. Their licensing guide can help determine the license of a dataset.
        \item For existing datasets that are re-packaged, both the original license and the license of the derived asset (if it has changed) should be provided.
        \item If this information is not available online, the authors are encouraged to reach out to the asset's creators.
    \end{itemize}

\item {\bf New assets}
    \item[] Question: Are new assets introduced in the paper well documented and is the documentation provided alongside the assets?
    \item[] Answer: \answerNA{}
    \item[] Justification: The paper does not introduce or release a new standalone dataset, pretrained model, or other reusable asset that requires the structured documentation described in this question. The distilled images generated by the proposed method are experimental outputs rather than a separately released dataset or model.

    \item[] Guidelines:
    \begin{itemize}
        \item The answer \answerNA{} means that the paper does not release new assets.
        \item Researchers should communicate the details of the dataset\slash code\slash model as part of their submissions via structured templates. This includes details about training, license, limitations, etc.
        \item The paper should discuss whether and how consent was obtained from people whose asset is used.
        \item At submission time, remember to anonymize your assets (if applicable). You can either create an anonymized URL or include an anonymized zip file.
    \end{itemize}

\item {\bf Crowdsourcing and research with human subjects}
    \item[] Question: For crowdsourcing experiments and research with human subjects, does the paper include the full text of instructions given to participants and screenshots, if applicable, as well as details about compensation (if any)?
    \item[] Answer: \answerNA{}
    \item[] Justification: The research does not involve crowdsourcing, human-subject experiments, participant recruitment, or collection of human-subject data.

    \item[] Guidelines:
    \begin{itemize}
        \item The answer \answerNA{} means that the paper does not involve crowdsourcing nor research with human subjects.
        \item Including this information in the supplemental material is fine, but if the main contribution of the paper involves human subjects, then as much detail as possible should be included in the main paper.
        \item According to the NeurIPS Code of Ethics, workers involved in data collection, curation, or other labor should be paid at least the minimum wage in the country of the data collector.
    \end{itemize}

\item {\bf Institutional review board (IRB) approvals or equivalent for research with human subjects}
    \item[] Question: Does the paper describe potential risks incurred by study participants, whether such risks were disclosed to the subjects, and whether Institutional Review Board (IRB) approvals (or an equivalent approval/review based on the requirements of your country or institution) were obtained?
    \item[] Answer: \answerNA{}
    \item[] Justification: The study does not involve human-subject research, participant recruitment, crowdsourcing, or collection of new data from human participants. Therefore, IRB approval or an equivalent review is not applicable.

    \item[] Guidelines:
    \begin{itemize}
        \item The answer \answerNA{} means that the paper does not involve crowdsourcing nor research with human subjects.
        \item Depending on the country in which research is conducted, IRB approval (or equivalent) may be required for any human subjects research. If you obtained IRB approval, you should clearly state this in the paper.
        \item We recognize that the procedures for this may vary significantly between institutions and locations, and we expect authors to adhere to the NeurIPS Code of Ethics and the guidelines for their institution.
        \item For initial submissions, do not include any information that would break anonymity (if applicable), such as the institution conducting the review.
    \end{itemize}

\item {\bf Declaration of LLM usage}
    \item[] Question: Does the paper describe the usage of LLMs if it is an important, original, or non-standard component of the core methods in this research? Note that if the LLM is used only for writing, editing, or formatting purposes and does \emph{not} impact the core methodology, scientific rigor, or originality of the research, declaration is not required.
    \item[] Answer: \answerNA{}
    \item[] Justification: The core methodology does not use an LLM as an important, original, or non-standard component. The proposed DeCO framework is based on teacher-guided discriminative evidence extraction and grid-based evidence composition for fine-grained dataset distillation.

    \item[] Guidelines:
    \begin{itemize}
        \item The answer \answerNA{} means that the core method development in this research does not involve LLMs as any important, original, or non-standard components.
        \item Please refer to our LLM policy in the NeurIPS handbook for what should or should not be described.
    \end{itemize}

\end{enumerate}

\end{document}